\documentclass[10pt,twocolumn,letterpaper]{article}

\usepackage{cvpr}
\usepackage{times}
\usepackage{epsfig}
\usepackage{graphicx}
\usepackage{amsmath}
\usepackage{amssymb}
\usepackage{multirow}
\usepackage{makecell}
\newtheorem{theorem}{Theorem}

\newtheorem{Proof}{Proof}
\usepackage{algorithm}
\usepackage{algorithmic}
\usepackage{color} 

\usepackage[breaklinks=true,bookmarks=false]{hyperref}

\cvprfinalcopy 

\def\cvprPaperID{****} 

\begin{document}

\title{Filter Grafting for Deep Neural Networks}

\author{Fanxu Meng$^{1*}$, Hao Cheng$^{2*\dagger}$, Ke Li$^2$, Zhixin Xu$^1$, Rongrong Ji$^{3,4}$, Xing Sun$^2$, Gaungming Lu $^{1\dagger}$\\ 
	$^1$  School of Computer Science and Technology, Harbin Institute of Technology (Shenzhen), China\\					
    $^2$ Tencent Youtu Lab, Shanghai, China \\
    $^3$ Department of Artificial Intelligence, School of Informatics, Xiamen University, China\\
    $^4$ Peng Cheng Laboratory, China\\
	{\tt \scriptsize \{louischeng, tristanli, winfredsun\}@tencent.com,   \{18S151514,xuzhixin\}@stu.hit.edu.cn, luguangm@hit.edu.cn, rrj@xmu.edu.cn}
	}

\maketitle

\begin{abstract}
	
	This paper proposes a new learning paradigm called \textbf{filter grafting}, which aims to improve the representation capability of Deep Neural Networks (DNNs). The motivation is that DNNs have unimportant (invalid) filters (e.g., $l_{1}$ norm close to 0). These filters limit the potential 
	of DNNs since they are identified as having little effect on the network. While filter pruning removes these invalid filters for efficiency consideration, filter grafting re-activates them from an accuracy boosting perspective. The activation is processed by grafting external information (weights) into invalid filters. To better perform the grafting process, we develop an \textbf{entropy-based criterion} to measure the information of filters and an \textbf{adaptive weighting strategy} for balancing the grafted information among networks. After the grafting operation, the network has very few invalid filters compared with its untouched state, enpowering the model with more representation capacity. We also perform extensive experiments on the classification and recognition tasks to show the superiority of our method. For example, the grafted MobileNetV2 outperforms the non-grafted MobileNetV2 by about 7 percent on CIFAR-100 dataset. Code is available at \textcolor{magenta}{https://github.com/fxmeng/filter-grafting.git}.

	
\end{abstract}

\section{Introduction}\label{sec_1}

\footnotetext[1] { In the author list, $^{\ast}$ denotes that authors contribute equally and are listed in alphabetical order; $^{\dagger}$ denotes corresponding authors.}
Since Krizhevsky \emph{et al}. \cite{krizhevsky2012imagenet} make a breakthrough in the 2012 ImageNet competition \cite{ILSVRC15}, researchers have got significant advancements in exploring various architectures for DNNs  (Simonyan \& Zisserman  \cite{Simonyan2014Very}; Szegedy \emph{et al}. \cite{Szegedy_2015_CVPR}; He \emph{et al}. \cite{he2016deep}; Lu \emph{et al}. \cite{lu2019super,lu2019sparse}). 
DNNs gradually become very popular and powerful models in areas including computer vision \cite{krizhevsky2012imagenet,lotter2016deep}, speech recognition \cite{graves2013speech}, and language processing \cite{zhang2015text}. However, recent studies show that DNNs have invalid (unimportant) filters \cite{li2016pruning}. These filters are identified as having a small effect on output accuracy. Removing certain filters could accelerate the inference of DNNs without hurting much performance. This discovery inspires many works studying how to decide which filters are unimportant \cite{molchanov2019importance} and how to effectively remove the filters with tolerable performance drop \cite{zhuo2018scsp,suau2018principal}.

\begin{figure}[t]
	\centering
	\includegraphics[width=8.5cm,height=5cm]{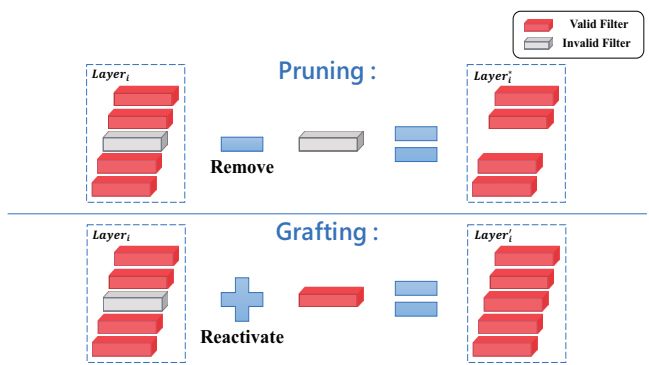}
	\caption{An illustration of the difference between filter pruning and filter grafting. For filter grafting, we graft external information into invalid filters without changing the model structure. (best viewed in color)}
	\label{figure: pruning&grafting}
\end{figure}


\begin{table*}[!t]
	\begin{center}
		\footnotesize
		\begin{tabular}{|c|c|c|c|c|}
			\hline
			methods& without changing model structure ? &one stage ? & without supervision ?   \\ 
			\hline
			filter pruning \cite{li2016pruning}  & $\times$ &$ \times$ & \checkmark
			\\
			\hline
			distillation \cite{hinton2015distilling}&\checkmark &$\times$ &$\times$
			\\	
			\hline	
			deep mutual learning \cite{zhang2018deep}&\checkmark &\checkmark &$\times$
			\\	
			\hline	
			RePr \cite{prakash2019repr} &$\checkmark$ &$\times$ &$\checkmark$
			\\
			\hline
			\textbf{filter grafting} &\checkmark &\checkmark &\checkmark
			\\
			\hline
		\end{tabular}
	\end{center}
	\caption{The difference between filter grafting and other learning methods}\label{Related_Work_Difference}
\end{table*}

However, it is unclear that whether directly abandoning such filters and components is the best choice.
What if, such traditional \emph{invalid} filters are indeed useful in certain senses?
The same story happens in the ensemble learning like boosting, where while a single weak classifier is poor, their combination and retraining might open a gate towards optimal performance.
Besides, given multiple networks, it is unclear whether one network can learn from the others.
In this paper, we investigate the possibility to re-activate the invalid filters in one network by bringing outside information.
This is achieved by proposing a novel filter grafting scheme, as illustrated in Figure \ref{figure: pruning&grafting}.
Filter grafting differs from filter pruning in the sense that we re-activate filters by assigning new weights, which maintains the number of layers and filters within each layer as the same. The grafted network has a higher representation capability since more valid filters in the network are involved in processing information. 

A key step in filter grafting is choosing proper information source (\emph{i.e.}, where should we graft the information from). In this paper, we thoroughly study this question and claim that we should graft the information from outside (other networks) rather than inside (self-network). 
Generally, we could train several networks in parallel. During training at certain epochs, we graft a network's meaningful filters into another network's invalid filters. By performing grafting, each network could learn external information from other networks. The details can be found in Section \ref{sec_3}.

There are three main contributions of this paper:

\begin{itemize}
	\item We propose a new learning paradigm called \textbf{filter grafting} for DNNs. Grafting could re-activate the invalid filters to improve the potential of DNNs without changing the network structure.
	
	\item An \textbf{entropy based criterion} and an \textbf{adaptive weighting strategy} are developed to further improve the performance of filter grafting method.
	
	\item We perform extensive experiments on classification and recognition tasks and show grafting could substantially improve the performance of DNNs. For example, the grafted MobileNetV2 achieves 78.32\% accuracies on CIFAR-100, which is about 7\% higher than non-grafted MobileNetV2. 
\end{itemize}

\section{Related Work}\label{sec_2}

\textbf{Filter Pruning.} 
Filter pruning aims to remove the invalid filters to accelerate the inference of the network. \cite{li2016pruning} first utilizes $l_{1}$ norm criterion to prune unimportant filters. Since then, more criterions came out to measure the importance of the filters. \cite{zhuo2018scsp} utilizes spectral clustering to decide which filter needs to be removed. \cite{suau2018principal} proposes an inherently data-driven method that utilizes Principal Component Analysis (PCA) to specify the proportion of the energy that should be preserved. \cite{wang2018exploring} applies subspace clustering to feature maps to eliminate the redundancy in convolutional filters. While instead of abandoning the invalid filters, filter grafting intends to activate them. It is worth noting that even though the motivation of filter grafting is opposite to pruning, grafting still involves choosing a proper criterion to decide which filters are unimportant. Thus different criterions from pruning are readily applied to grafting.

\textbf{Distillation and Mutual Learning.} Grafting may involve training multiple networks in parallel. Thus this process is similar to distillation \cite{hinton2015distilling} and mutual learning \cite{zhang2018deep}. The difference between grafting and distillation is that distillation is a `two-stage' process. First, we need to train a large model (teacher), then use the trained model to teach a small model (student). While grafting is a `one-stage' process, we graft the weight during the training process. The difference between mutual learning and grafting is that mutual learning needs a mutual loss to supervise each network to learn and do not generalize well to multiple networks. While grafting does not need supervised loss and performs much better when we add more networks into the training process. Also, we graft the weight at each epoch instead of each iteration, thus greatly reduce communication costs among networks.

\textbf{RePr.} RePr \cite{prakash2019repr} is similar to our work which considers improving network on the filter level. However, the motivation of RePr is that there exists unnecessary overlaps in the features captured by the network’s filters. RePr first prunes overlapped filters to train the sub-network, then restores the pruned filters and re-trains the full network. In this sense, RePr is a multi-stage training algorithm. In contrast, the motivation of filter grafting is that the filter whose $l_{1}$ norm is smaller contributes less to the network output. Thus the filters that each method operates are different. Also grafting is a one-stage training algorithm which is more efficient. To better illustrate how grafting differs from the above learning types. We draw a table in Table \ref{Related_Work_Difference}. From Table \ref{Related_Work_Difference}, filter grafting is a one stage learning method, without changing network structure and does not need supervised loss.

\section{Filter Grafting}\label{sec_3}
This section arranges as follows: In Section \ref{sec_3_1}, we study the source of information that we need during grafting process; In Section \ref{sec_3_2}, we propose two criterions to calculate the information of filters; In Section \ref{sec_3_3}, we discuss how to effectively use the information for grafting; In Section \ref{sec_3_4}, we extend grafting method to multiple networks and propose our final entropy-based grafting algorithm. 

\subsection{Information Source for Grafting}\label{sec_3_1}
In the remaining, we would call the original invalid filters as 'rootstocks' and call the meaningful filters or information to be grafted as 'scions', which is consistent with botany interpretation for grafting. Filter grafting aims to transfer information (weights) from scions to rootstocks, thus selecting useful information is essential for grafting. In this paper, we propose three ways to get scions.

\subsubsection{Noise as Scions}\label{sec_3_1_1}

A simple way is to graft gaussian noise $ \mathcal{N}(0,\sigma_{t})$ into invalid filters, since gaussian noise is commonly used for weight initialization of DNNs \cite{kumar2017weight,he2015delving}. 
Before grafting, the invalid filters have smaller $l_{1}$ norm and have little effects for the output.
But after grafting, the invalid filters have larger $l_{1}$ norm and begin to make more effects to DNNs. 

\begin{equation}\label{noise_decrease}
\sigma_{t}=a^t(0<a<1)
\end{equation}

We also let $\sigma_{t}$ decrease over time (see \eqref{noise_decrease}), since too much noise may make the model harder to converge. 

\subsubsection{Internal Filters as Scions}\label{sec_3_1_2}

\begin{figure}[!h]
	\centering
	\includegraphics[width=8.5cm,height=5cm]{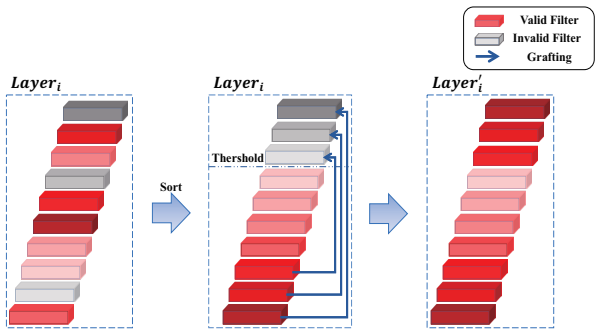}
	\caption{Grafting internal filters. We first sort the filters by $l_{1}$ norm, then graft the weights from filters with larger $l_{1}$ norm into filters with smaller $l_{1}$ norm. (best viewed in color)}
	\label{figure: inside_grafting}
\end{figure}

Instead of adding random noise, we add the weights of other filters ($l_{1}$ norm is bigger) into the invalid filters ($l_{1}$ norm is smaller). Grafting is processed inside a single network. Specifically, for each layer, we sort the filters by $l_{1}$ norm and set a threshold $\gamma$. For filters whose $l_{1}$ norm are smaller than $\gamma$, we treat these filters as invalid ones. Then we graft the weights of the $i$-th largest filter into the $i$-th smallest filter. This procedure is illustrated in Figure \ref{figure: inside_grafting}.

Since the invalid filters have new weights with larger $l_{1}$ norm, they can be activated to have a bigger influence on the output. But this method does not bring new information to the network since the weights are grafted inside the self network. We further evaluate it via the language of information theory. To simplify the proving process, we deal with two filters in a certain layer of the network (See Theorem \ref{theo1}, proof can be found in the supplementary material). From Theorem \ref{theo1}, selecting internal filters as scions does not bring new information. The experiment in Section \ref{sec_4_1} is also consistent with our analysis.

\begin{theorem}\label{theo1}
	Suppose there are two filters in a certain layer of the network, denoted as random variables $X$ and $Y$. $Z$ is another variable which satisfies $Z = X+Y$, then $H(X,Y) = H(X,Z) = H(Y,Z)$, where $H$ denotes the entropy from information theory.
\end{theorem}

\subsubsection{External Filters as Scions}	\label{sec_3_1_3}

In response to the shortcomings of adding random noise and weights inside a single network, we select external filters from other networks as scions.
Specifically, we could train two networks, denoted as $M_{1}$ and $M_{2}$, in parallel. During training at certain epochs, we graft the valid filters' weights of $M_{1}$ into the invalid filters of $M_{2}$. Compared to the grafting process in Section \ref{sec_3_1_2}, we make two modifications:
\begin{itemize}
	\item The grafting is processed at layer level instead of filter level, which means we graft the weights of all the filters in a certain layer in $M_{1}$ into the same layer in $M_{2}$ (also $M_{2}$ into $M_{1}$, inversely). Since two networks are initialized with different weights, the location of invalid filters are statistically different and only grafting information into part of filters in a layer may break layer consistency (see more analyses and experimental results in the supplementary material).  By performing grafting, the invalid filters of two networks can learn mutual information from each other.
	\item When performing grafting, the inherent information and the extoic information are weighted. Specifically, We use $W_{i}^{M_{2}}$ denotes the weights of the $i$-th layer of $M_{2}$, $W_{i}^{M_{2}^{'}}$ denotes the weights of the $i$-th layer of $M_{2}$ after grafting. Then:
	\begin{equation}\label{weighting}
	W_{i}^{M_{2}^{'}} = \alpha W_{i}^{M_{2}} + (1-\alpha) W_{i}^{M_{1}}~~~ (0<\alpha<1)
	\end{equation}
	Suppose $W_{i}^{M_{2}}$ is more informative than $W_{i}^{M_{1}}$, then $\alpha$ should be larger than $0.5$. 
\end{itemize}

\begin{figure}[!h]
	\centering
	\includegraphics[width=8.5cm,height=6cm]{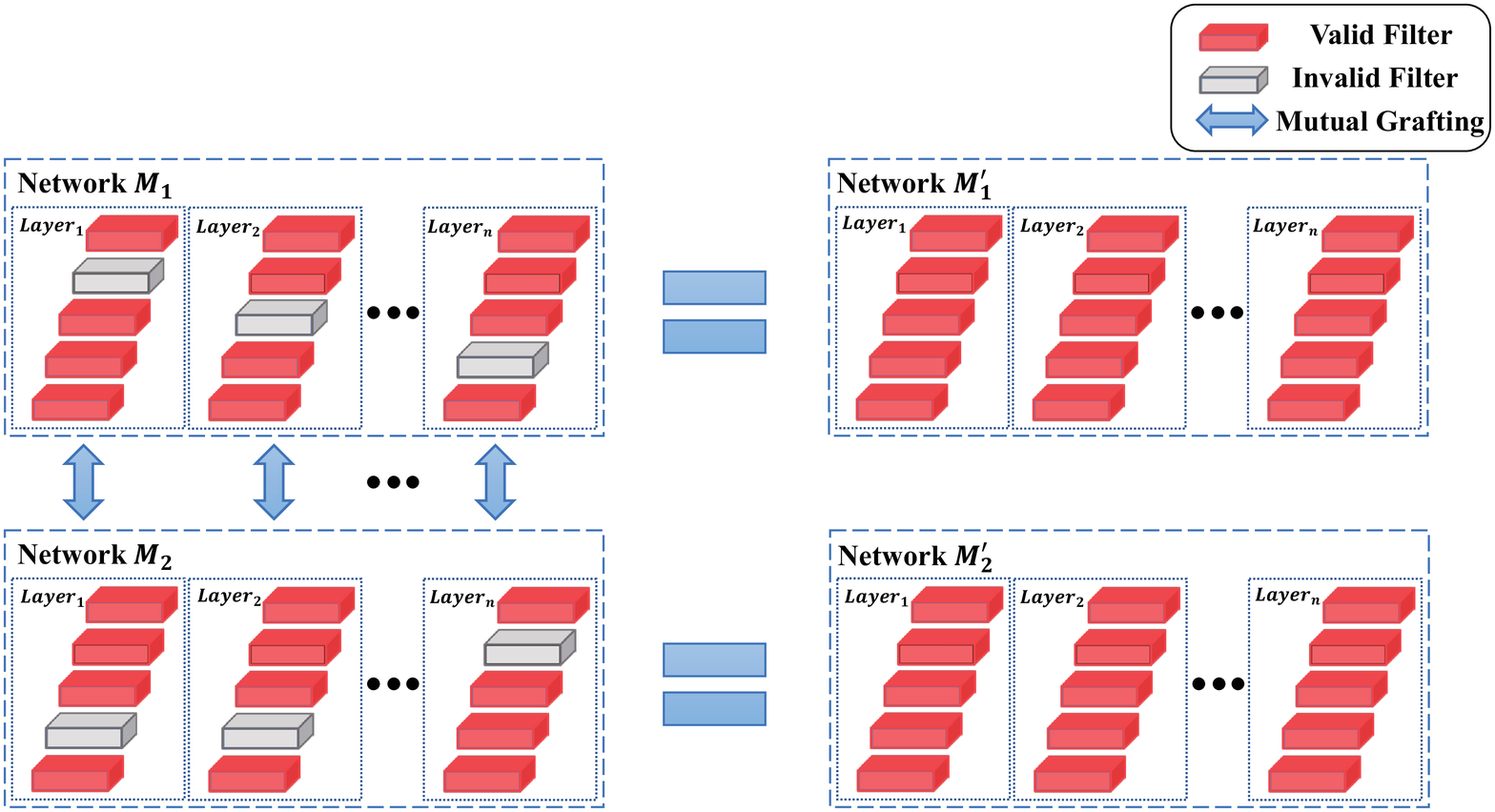}
	\caption{Grafting between two networks. Each network accepts information from the other network. (best viewed in color)}
	\label{figure: mutual_grafting}
\end{figure}

The two networks grafting procedure is illustrated in Figure \ref{figure: mutual_grafting}. From Equation \eqref{weighting} and Figure \ref{figure: mutual_grafting}, there are two key points in grafting: 1) how to calculate the information of $W_{i}^{M_{1}}$ and $W_{i}^{M_{2}}$; 2) How to decide the weighting coefficient $\alpha$.
We thoroughly study these two problems in Section \ref{sec_3_2} and Section \ref{sec_3_3}. Also, we hope to increase the diversity of two networks, thus two networks are initialized differently and some hyper-parameters of two networks are also different from each other (e.g., learning rate, sampling order of data \dots).  It is worth noting that when performing grafting algorithm on two networks, the two networks have the same weights after grafting process from \eqref{weighting}. But grafting is only performed at each epoch. For other iteration steps, since the two networks are learned with different hyper-parameters, their weights are different from each other. Also, this problem disappears when we add more networks ($N>2$) in grafting algorithm. Multiple networks grafting can be found in Section \ref{sec_3_4}.

\subsection{Criterions for Calculating Information of Filters and Layers}\label{sec_3_2}		
In this section, we study two criterions to calculate the information of filters or layers.

\subsubsection{$L_{1}$ norm }\label{sec_3_2_1}	
In previous sections, we use $l_{1}$ norm to measure the information of filters. Denote $W_{i,j}\in \mathbb{R}^{N_{i}\times K\times K }$ as the weight of the $j$-th filter in the $i$-th convolutional layer, where $N_{i}$ is the number of filters in $i$-th layer. Its $l_{1}$ norm can be presented by:
\begin{equation}\label{l1 norm}
\|W_{i,j}\|_{1}= \sum_{n=1}^{N_{i}}\sum_{k_{1}=1}^{K}\sum_{k_{2}=1}^{K}|W_{i,j}(n,k_{1},k_{2})|
\end{equation}

The $l_{1}$ norm criterion is commonly used in many research \cite{li2016pruning,Ye2018Rethinking,Yang2018Soft}. But recent studies show smaller-norm-less-important criterion is not always true. One special case is that 0-1 regularly arranged filters are better than all 1 filters. \cite{He_2019_CVPR} also points out that there are some pre-requisites to utilize this “smaller-norm-less-important” criterion. Otherwise, pruning
may hurt valid filters.

\begin{algorithm*}[!t]
	\caption{Entropy-based Multiple Networks Grafting}
	\label{algorithm2}
	\begin{algorithmic}[2]
		\renewcommand{\algorithmicrequire}{\textbf{Input:}}
		\renewcommand{\algorithmicensure}{\textbf{Iteration:}}
		\REQUIRE ~~\\
		Number of networks $K$, $M_{k}$ denotes the $k$-th network; Number of layers $L$; 
		Training iterations $\mathcal N = \{1,\ldots, N_{max}  \}$; Number of iterations for each epoch $N_{T}$; Training dataset $\mathcal D$;
		Initial weights for each layer of each network $\{\mathbf{W}_{l}^{M_{k}}:k=1,\ldots, K; l=1,\ldots, L \}$; Different hyper-parameters for each network $\{\mathbf{\lambda}_k:k=1,\ldots K\}$.	\ENSURE ~~\\
		\textbf{for} $n = 1$ to $N_{max}$   
		\\
		\qquad\textbf{for} $\mathbf{k}\in\{1,\ldots K\}$, $\mathbf{l}\in\{1,\ldots L\}$ \textbf{parallel do}
		\\
		\qquad\qquad Update model parameters $\mathbf{W}^{M_{k}}_{l}$ based on $\mathcal D$ with $\mathbf{\lambda}_k$ ~~~//Update model weights at each iteration.
		\\
		\qquad\qquad \textbf{if} $n~mod~N_{T} = 0$ 
		\\
		\qquad\qquad\qquad Get the weighting coefficient $\alpha$ from \eqref{coefficient} ~~~~~~~~//Graft model weights at each epoch.
		\\
		\qquad\qquad\qquad  $\mathbf{W}^{M_{k}}_{l} = \alpha \mathbf{W}^{M_{k}}_{l} + (1-\alpha)\mathbf{W}^{M_{k-1}}_{l}$  
		\\
		\qquad\qquad \textbf{end if}
		\\
		\qquad\textbf{end for}
		\\
		\textbf{end for}
		\\
	\end{algorithmic}
\end{algorithm*}

\subsubsection{Entropy}	\label{sec_3_2_2}

While $l_{1}$ norm criterion only concentrates on the absolute value of filter's weight, we pay more attention to the variation of the weight. A problem of $l_{1}$ norm criterion is that $l_{1}$ norm neglects the variation of the weight. Suppose a filter's weight $W_{i,j}\in \mathbb{R}^{N_{i}\times K\times K }$  satisfies $W_{i,j}(n,k_{1},k_{2}) = a$ for each $n\in \{1,\dots,N_{i}\} $ and $k_{1},k_{2}\in \{1,\dots,K\}$, Each single value in $W_{i,j}$ will be the same. Thus when using $W_{i,j}$ to operate convolution on the input, each part of the input is contributed equally to the output even though $a$ is big. Thus the filter can not discriminate which part of the input is more important. Based on the above analyses, we choose to measure the variation of the weight. We suppose each value of $W_{i,j}$ is sampled from a distribution of a random variable $X$ and use the entropy to measure the distribution. Suppose the distribution satisfies $P(X=a)=1$, then each single value in $W_{i,j}$ is the same and the entropy is 0. 
While calculating the entropy of continuous distribution is hard, we follow the strategy from \cite{shwartz2017opening,cheng2019utilizing}. We first convert continuous distribution to discrete distribution. Specifically, we divide the range of values into $m$ different bins and calculate the probability of each bin. Finally, the entropy of the variable can be calculated as follows:
\begin{equation}\label{binning}
H(W_{i,j}) =-\sum_{k=1}^{B}p_k\log p_k
\end{equation}
Where $B$ is the number of bins and $p_{k}$ is the probability of bin $k$. A smaller score of $H(W_{i,j})$ means the filter has less variation (information).

Suppose layer $i$ has $C$ filters, then the total information of the layer $i$ is:
\begin{equation}\label{info_layer_pre}
H(W_{i}) = \sum_{j=1}^{C} H_{i,j}
\end{equation}
But one problem of \eqref{info_layer_pre} is that it neglects the correlations among the filters since \eqref{info_layer_pre} calculates each filter's information independently. To keep layer consistency, we directly calculate the entropy of the whole layer's weight $W_{i}\in \mathbb{R}^{N_{i}\times N_{i+1}\times K\times K }$ as follows:
\begin{equation}\label{info_layer}
H(W_{i})=-\sum_{k=1}^{B}p_k\log p_k
\end{equation}
Different from \eqref{binning}, the values to be binned in \eqref{info_layer} are from the weight of the whole layer instead of a single filter. In the supplementary material, we prove layer consistency is essential for grafting algorithm.

\subsection{Adaptive Weighting in Grafting}\label{sec_3_3}
\begin{figure}[!h]
	\centering
	\includegraphics[width=8cm,height=5cm]{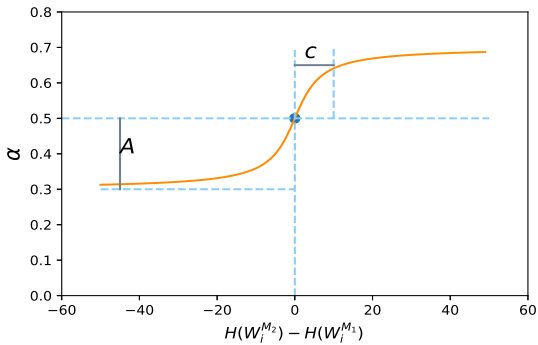}
	\caption{The adaptive coefficient in grafting process.}
	\label{figure: adaptive_w}
\end{figure}

In this part, we propose an adaptive weighting strategy for weighting two models' weight from \eqref{weighting}.  Denote $W_{i}^{M_{1}}$ and $H(W_{i}^{M_{1}})$ as the weight and information of layer $i$  in network $M_{1}$, respectively. The calculation of $H(W_{i}^{M_{1}})$ can be referred to \eqref{info_layer}. We enumerate two conditions that need to be met for calculating the coefficient $\alpha$.
\begin{itemize}
	\item The coefficient $\alpha$ from \eqref{weighting} should be equal to 0.5 if $H(W_{i}^{M_{2}}) = H(W_{i}^{M_{1}})$ and 
	larger than 0.5 if $H(W_{i}^{M_{2}}) > H(W_{i}^{M_{1}})$.
	\item Each network should contain part of self information even though $H(W_{i}^{M_{2}}) \gg H(W_{i}^{M_{1}})$ or $H(W_{i}^{M_{2}}) \ll H(W_{i}^{M_{1}})$.
\end{itemize}
In response to the above requirements, the following adaptive coefficient is designed: 
\begin{equation}\label{coefficient}
\alpha=A\ast(arctan(c\ast(H(W_{i}^{M_{2}})-H(W_{i}^{M_{1}}))))+0.5
\end{equation}
where $A$ and $c$ from \eqref{coefficient} are the fixed hyper-parameters. $\alpha$ is the coefficient of \eqref{weighting}. We further depict a picture in 
Figure \ref{figure: adaptive_w}. We can see this function well satisfies the above conditions.

\subsection{Extending Grafting to Multiple Networks}\label{sec_3_4}

\begin{figure}[!h]
	\centering
	\includegraphics[width=8.5cm,height=5cm]{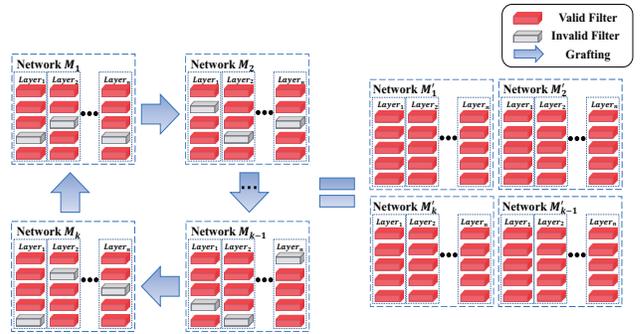}
	\caption{Grafting with multiple networks. The network $M_{k}$ accepts information from $M_{k-1}$. (best viewed in color)}
	\label{figure: multiple_grafting}
\end{figure}

Grafting method can be easily extended to a multi-networks case, as illustrated in Figure \ref{figure: multiple_grafting}. At each epoch during training, each network $M_{k}$ accepts the information from $M_{k-1}$. After certain training epochs, each network contains information from all the other networks. The weighting coefficient is also calculated adaptively. From Section \ref{sec_4_5}, we find that by using grafting to train multiple networks, each network achieves much performance gain. 
We propose our entropy-based grafting in Algorithm \ref{algorithm2}. It is worth noting that grafting is performed on multiple networks in parallel, which means when we use $\mathbf{W}^{M_{k-1}}_{l}$ to update $\mathbf{W}^{M_{k}}_{l}$, $\mathbf{W}^{M_{k-1}}_{l}$ has not been updated by grafting yet.

\section{Experiment}\label{sec_4}
This section arranges as follows: In Section \ref{sec_4_1}, we examine how different information sources affect grafting method; In Section \ref{sec_4_2}, we prove entropy-based grafting is better than $l_{1}$ norm-based grafting; In Section \ref{sec_4_3}, we analyze the training diversity when performing grafting; In Section \ref{sec_4_4}, we compare grafting with other learning methods; In Section \ref{sec_4_5}, we show by using multiple-networks, grafting could greatly improve the performance of the network; In Section \ref{sec_4_6} and Section \ref{sec_4_7}, we examine grafting on close-set classification and open-set recognition tasks; In Section \ref{sec_4_8}, we further analyze the effectiveness of grafting algorithm. All the experiments are reproducible. The code is available upon requirement and will be released online. 

\subsection{Selecting Useful Information Source}\label{sec_4_1}

We propose three ways to get scions in Section \ref{sec_3} and experimentally examine the three ways on CIFAR-10 and CIFAR-100 datasets in Table \ref{table:branches}. Vanilla DNN training without grafting is taken as the baseline. All the methods use MobileNetV2 as the base model. For a fair comparison, the same hyper-parameters are deployed for each method: mini-batch size (256), optimizer (SGD), initial learning rate (0.1), momentum (0.9), weight decay (0.0005), number of epochs (200), learning rate decay (0.1 at every 60 epochs).
'External' here involves training two networks in parallel. In practice, we find the performance of each network in the 'external' method is very close to each other. Thus in the remaining, we always record the first network's performance.

\begin{table}[!h]
	\begin{center}
		\begin{tabular}{|c|c|c|} 
			\hline 
			&CIFAR-10 &CIFAR-100\\ 
			\hline
			baseline&92.42&71.44\\
			noise&92.51&72.34\\
			internal&92.68&72.38\\
			external&\textbf{92.94}&\textbf{72.90}\\
			\hline
		\end{tabular}\\
	\end{center}
	\caption{Comparison of different scion sources. }
	\label{table:branches}
\end{table}

From Table \ref{table:branches}, the performance of `internal scions' is similar to 'noise', since we prove in Theorem \ref{theo1} that choosing internal filters as scions does not bring new information to the network. While choosing external filters as scions achieves the best result among the three methods. In the remaining, all the grafting experiments choose external filters as scions.

\subsection{Comparison of $L_{1}$ norm \& Entropy Criterions}\label{sec_4_2}

We propose two criterions to measure the inherent information of filters in Section \ref{sec_3_2}. In this part, we quantitatively evaluate the $l_{1}$ norm-based grafting and the entropy-based grafting on CIFAR-10 and CIFAR-100 dataset. The results are listed in Table \ref{table:norm_entropy}. Two networks are used for grafting, with an identical model structure and training hyper-parameters. From Table \ref{table:norm_entropy}, we can find that, entropy-based grafting beats $l_{1}$ norm-based grafting on every model and dataset setting.

\begin{table}[!h]
	\begin{center}
		\begin{tabular}{|c|c|c|c|} 
			\hline 
			model&method&CIFAR-10 &CIFAR-100\\ 
			\hline
			ResNet32&baseline&92.83&69.82\\
			&$l_{1}$ norm&93.24&70.69\\
			&entropy&\textbf{93.33}&\textbf{71.16}\\
			\hline
			ResNet56&baseline&93.50&71.55\\
			&$l_{1}$ norm&94.09&72.73\\
			&entropy&\textbf{94.28}&\textbf{73.09}\\
			\hline
			ResNet110&baseline&93.81&73.21\\
			&$l_{1}$ norm&94.37&73.65\\
			&entropy&\textbf{94.60}&\textbf{74.70}\\
			\hline
			MobileNetV2&baseline&92.42&71.44\\
			&$l_{1}$ norm&92.94&72.90\\
			&entropy&\textbf{93.53}&\textbf{73.26}\\
			\hline
		\end{tabular}
	\end{center}
	\caption{Comparison of grafting by $l_{1}$ norm \& entropy. }
	\label{table:norm_entropy}
\end{table}

\subsection{Evaluation of Training Diversity in Grafting}\label{sec_4_3}

We find that the performance raises when we increase the training diversity of two networks. Since grafting is about transferring weights between models, the network can learn better if the external information (weights) has more variations. 
To achieve this, we could diversify the hyper-parameters setting (sampling order and learning rate in our case) to see how these factors affect grafting performance. 
The results are listed in Table \ref{table:diversity}. Cosine annealing LR schedule with different initial learning rate is set for each model in different LR case (This ensures that at each step, the learning rate for each model is different).
We find that the weight variations brought by  sampling order and learning rate enrich the grafting information and thus encourage the models to learn better. In the remaining, when performing grafting, all the networks use different hyper-parameters in terms of data loader and learning rate.

\begin{table}[!h]
	\begin{center}
		\begin{tabular}{|c|c|c|c|c|c|} 
			\hline 
			different order&different LR&CIFAR10 &CIFAR100\\ 
			\hline
			$\times$&$\times$&93.05&71.91\\
			$\checkmark$&$\times$&93.53&73.26\\
			$\checkmark$&$\checkmark$&\textbf{94.20}&\textbf{74.15}\\
			\hline
		\end{tabular}
	\end{center}
	\caption{Hyper-parameters verification for grafting. The backbone is MobileNetV2. }
	\label{table:diversity}
\end{table}


\begin{table*}[!t]
	\begin{center}
		\begin{tabular}{c|c|c|c|c|c|c} 
			\hline 
			Dataset&method&ResNet32&ResNet56&ResNet110&MobileNetV2&WRN28-10\\
			\hline 
			&baseline&92.83&93.50&93.81&92.42&95.75\\
			CIFAR-10&distillation \cite{hinton2015distilling}&93.11&92.05&92.34&92.37&95.70\\
			&mutual learning \cite{zhang2018deep}&92.80&--&--&--&95.66\\
			&RePr \cite{prakash2019repr}&93.90&--&94.60&--&--\\
			&filter grafting&\textbf{93.94}&\textbf{94.73}&\textbf{94.96}&\textbf{94.20}&\textbf{96.40}\\
			\hline 
			&baseline&69.82&71.55&73.21&71.44&80.65\\
			CIFAR-100&distillation \cite{hinton2015distilling}&70.96&72.03&73.32&73.37&81.03\\
			&mutual learning \cite{zhang2018deep}&70.19&--&--&--&80.28\\
			&RePr \cite{prakash2019repr}&69.90&--&73.60&--&--\\
			&filter grafting&\textbf{71.28}&\textbf{72.83}&\textbf{75.27}&\textbf{74.15}&\textbf{81.62}\\
			\hline 
		\end{tabular}
	\end{center}
	\caption{Comparion of filter grafting with other learning methods. `--' denotes the result is not reported in the corresponding paper.}
	\label{table:cifar}
\end{table*}
\subsection{Comparing Grafting with Other Methods}\label{sec_4_4}

We thoroughly study the difference between grafting and other learning methods in Table \ref{Related_Work_Difference}. 
In this part, we experimentally compare grafting with other methods on CIFAR-10 and CIFAR-100 datasets in Table \ref{table:cifar}. 

For a fair comparison, `distillation', `mutual learning' and `filter grafting' all involve training two networks. The difference between distillation and grafting is that distillation is a two-stage training procedure. When performing distillation, we first train one network until convergence, then we use the network, as a teacher, to distill knowledge into the student network. For a fair comparison with grafting, the network structrue for teacher and student is the same which is consistent with the setting in \cite{zhang2018deep}. While for grating, training is completed in one stage without the retraining process. The difference between mutual learning and grafting is that mutual learning trains two networks with another strong supervised loss and communication costs are heavy between networks. One should carefully choose the coefficient for mutual supervised loss and main loss when using the mutual learning method. While for grafting, transferring weights does not need supervision. We graft the weights by utilizing entropy to adaptively calculate the weighting coefficient which is more efficient. The results from Table \ref{table:cifar} show that filter grafting achieves the best results among all the learning methods.

\subsection{Grafting with Multiple Networks}\label{sec_4_5}

The power of filter grafting is that we could greatly increase the performance by involving more networks in grafting algorithm. We examine the effect of multi-networks grafting in Table \ref{table:multiple models}. 

\begin{table}[!h]
	\begin{center}
		\begin{tabular}{|c|c|c|} 
			\hline 
			method&CIFAR-10 &CIFAR-100\\ 
			\hline
			baseline& 92.42 & 71.44 \\
			\hline
			2 models grafting&94.20&74.15\\
			3 models grafting&94.55&76.21\\
			4 models grafting&95.23&77.08\\
			6 models grafting&\textbf{95.33}&\textbf{78.32}\\
			8 models grafting&95.20&77.76\\
			\hline
			6 models ensemble&94.09&76.75\\
			\hline
		\end{tabular}\\
	\end{center}
	\caption{Grafting with multiple networks. The base netwok is MobileNetV2.}
	\label{table:multiple models}
\end{table}

As we raise the number of networks, the performance gets better. For example, the performance with 6 models grafting could outperform the baseline by about 7 percent which is a big improvement. The reason is that MobileNetV2 is based on depth separable convolutions, thus the filters may learn insufficient
knowledges. Filter grafing could help filters learn complementary knowledges from other networks, which greatly improves the network's potential. Also it is worth noting that the result of 6 models grafting is even better than 6 models ensembles. But unlike ensemble, grafting only maintains one network for testing. However, the performance stagnates when we add the number of models to 8 in grafting algorithm. We assume the cause might be that the network receives too much information from outside which may affect its self-information for learning. How to well explain this phenomenon is an interesting future work.

\subsection{Grafting on ImageNet}\label{sec_4_6}

To test the performance of grafting on a larger dataset, we also validate grafting on ImageNet, an image classification dataset with over 14 million images. We compare grafting with the baseline on ResNet18 and ResNet34 models. The baseline hyper-parameters' setting is consistent with official PyTorch setting for ImageNet\footnote{https://github.com/pytorch/examples/tree/master/imagenet}: minibatch size (256), initial learning rate (0.1), learning rate decay (0.1 at every 30 epochs), momentum (0.9), weight decay (0.0001), number of epochs (90) and optimizer (SGD). To increase the training diversity, we use different learning rates and data loaders for two networks when performing grafting. The other hyper-parameters' setting is consistent with the baseline. The results are shown in Table \ref{table:Imagenet}.

\begin{table}[!h]
	\begin{center}
		\begin{tabular}{|c|c|c|c|} 
			\hline 
			model&method&top-1&top-5\\ 
			\hline
			ResNet18&baseline&69.15&88.87\\
			&grafting&\textbf{71.19}&\textbf{90.01}\\
			\hline
			ResNet34&baseline&72.60&90.91\\
			&grafting&\textbf{74.58}&\textbf{92.05}\\
			\hline
		\end{tabular}\\
	\end{center}
	\caption{Grafting on ImageNet Dataset}
	\label{table:Imagenet}
\end{table}

From Table \ref{table:Imagenet}, we can find grafting performs better than the baseline. Thus grafting also can handle larger datasets.

\subsection{Grafting on ReID Task}\label{sec_4_7}

Grafting is a general training method for convolutional neural networks. Thus grafting can not only apply to the classification task but also other computer vision tasks. In this part, we evaluate the grafting on Person re-identification (ReID) task, an open set retrieval problem in distributed multi-camera surveillance, aiming to match people appearing in different non-overlapping camera views. We conduct experiments on two person ReID datasets: Market1501 \cite{market1505} and DukeMTMC-ReID (Duke) \cite{Ristani2016Performance,zheng2017unlabeled}.
The baseline hyper-parameters' setting is consistent with \cite{zhou2019osnet}: mini-batch size (32), pretrained (True), optimizer (amsgrad), initial learning rate (0.1), learning rate decay (0.1 at every 20 epochs), number of epochs (60). Besides data loaders and learning rate, the other hyper-parameters' setting is consistent with the baseline.The result is shown in Table \ref{table:ReID}.

\begin{table}[!h]
	\begin{center}
		\small
		\begin{tabular}{|c|c|cc|cc|} 
			\hline 
			model &method&\multicolumn{2}{c|}{Market1501} &\multicolumn{2}{c|}{Duke}\\ 
			&&mAP&rank1&mAP&rank1\\ 
			\hline
			ResNet50&baseline&67.6&86.7&56.2&76.2\\
			&2 models&70.6&87.8&60.8&79.8\\
			&4 models&\textbf{73.33}&\textbf{89.2}&\textbf{62.1}&\textbf{79.8}\\
			\hline
			MobileNetV2&baseline&56.8&81.3&47.6&71.7\\
			&2 models&63.7&85.2&53.4&76.1\\
			&4 models&\textbf{64.5}&\textbf{85.8}&\textbf{54.3}&\textbf{76.3}\\
			\hline
		\end{tabular}\\
	\end{center}
	\caption{Grafting on ReID Task}
	\label{table:ReID}
\end{table}

From table \ref{table:ReID}, for each model and each dataset, grafting performs better than the baseline. Besides, as mentioned before, increasing the number of networks in grafting can further improve the performance.

\subsection{Effectiveness of Grafting}\label{sec_4_8}

\begin{figure}[!h]
	\centering
	\includegraphics[width=7cm,height=4.5cm]{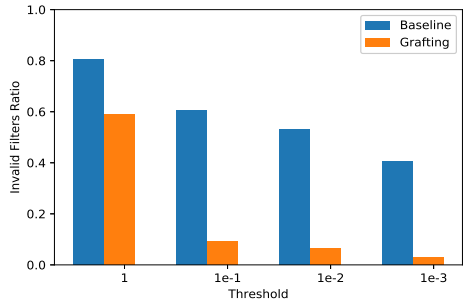}
	\caption{Ratio of filters whose $l_{1}$ norm under some threshold. }
	\label{figure: l1_threshold}
\end{figure}

\begin{figure}[!h]
	\centering
	\includegraphics[width=7cm,height=4.5cm]{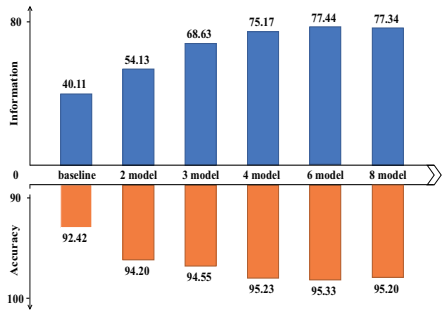}
	\caption{ This figure depicts the entropy and accuracy of the baseline network and grafted network. The network's information is defined as the sum of all the layers' entropy in a \textbf{single} network. The $x$ axis denotes the number of networks parallelly trained in grafting algorithm. }
	\label{figure: entropy_compare}
\end{figure}

In this part, we further analyze the effectiveness of the grafting method. To prove grafting does improve the potential of the network, we calculate the number of invalid filters and information gain after the training process. We select MobileNetV2, which is trained on CIFAR-10 with grafting algorithm, for this experiment. The same network structure without grafting is chosen as the baseline. Experimental results are reported in Figure \ref{figure: l1_threshold} and Figure \ref{figure: entropy_compare}. 

From Figure \ref{figure: l1_threshold}, under the threshold of 1e-3, there are about 50\% filters are invalid or unimportant for the base network, whereas the grafted network only has a small part of filters counted as `invalid', which shows grafting does help network reduce invalid filters.
From Figure \ref{figure: entropy_compare}, the model trained by grafting contains more information than the baseline. Also, the network can gain more information by training multiple networks for grafting method.  Thus from the above analysis, we confirm that grafting could improve the potential of neural networks. 
More analyses can be found in the supplementary material, including the evaluation of invalid filters' locations, necessity of keeping layer consistency and efficiency of adaptive weighting strategy.

\section{Conclusion and Discussion}\label{sec_5}
In this work, a new learning paradigm called `\textbf{filter grafting}' is proposed. We argue that there are two key points for effectively applying filter grafting algorithm: 1) How to choose proper criterion to calculate the inherent information of filters in DNNs. 2) How to balance the coefficients of information among networks. To deal with these two problems, we propose \textbf{entropy-based criterion} and \textbf{adaptive weighting strategy} to increase the network's performance. But this is not the only solution. Other criterions or methods could be developed to improve the grafting algorithm further. Heuristically, there are some future directions to be considered: 1) How to improve the network's performance with larger number of networks in grafting algorithm; 2) How to apply grafting on multiple networks with different network structures.



\newpage

{\small
	\bibliographystyle{ieee_fullname}
	\bibliography{egbib}
}

\newpage

\section{Supplementary Material}
This is the supplementary material for the paper "Filter Grafting for Deep Neural Networks". Section \ref{sec_s_2} proves  the locations of invalid filters 
are statistically different among networks. Section \ref{sec_s_3} shows layer consistency is essential for grafting algorithm. Section \ref{sec_s_4} further proves to keep layer consistency, the layer's information should be calculated from Equation (6) rather than Equation (5) from the main paper. Section \ref{sec_s_5} compares adaptive weighting strategy with fixed weighting strategy for grafting algorithm. Section \ref{sec_s_1} proves the Theorem \ref{theo1} from Section \ref{sec_3_1_2}.

\subsection{Locations of Invalid Filters}\label{sec_s_2}
We mentioned in Section 3.1.3 from the main paper that since two networks are initialized with different weights, the locations of invalid filters are statistically different. In this part, we perform an experiment to verify our claim. Specifically, we parallelly train two networks with the same structure and record the invalid filters in each layer of each network (20\% filters are counted as `invalid' in each layer). Then by calculating IoU (Intersection over Union) for the positions of invalid filters, we could verify our statement. A small IoU means that the locations of invalid filters are mostly different between two networks.

\begin{table}[!h]
	\begin{center}
		\begin{tabular}{|c|c|c|c|} 
			\hline
			model&layer-5&layer-10&layer-15\\
			\hline
			ResNet32&0.00&0.00&0.20\\
			\hline
			MobileNetV2&0.05&0.14&0.17\\
			\hline
		\end{tabular}\\
	\end{center}
	\caption{IoU for invalid filters' location.}
	\label{table:IoU}
\end{table}

From Table \ref{table:IoU}, the results have proved that the locations of invalid filters are statistically different between networks. Thus there exists little chance that the weight of an invalid filter is grafted into another invalid filter.

\subsection{Layer Consistency}\label{sec_s_3}
In Section 3.1.3 of the paper, we mentioned that to keep layer consistency, we should graft the weight in layer level instead of filter level.
We perform an experiment on two networks $M_{1}$ and $M_{2}$ to verify our claim. For filter level grafting, we sort filters by entropy in $M_{2}$ to get the invalid ones, and graft corresponding filters from $M_{1}$ into $M_{2}$.
To get a fair comparison, hyper-parameters are equally deployed for two methods. From Table \ref{table:filter level grafting}, layer level grafting performs better than filter level grafting.

\begin{table}[!h]
	\begin{center}
		\begin{tabular}{|c|c|c|c|} 
			\hline
			model&method&CIFAR-10&CIFAR-100\\
			\hline
			ResNet32&filter level&93.49&70.79\\
			&layer level&\textbf{93.94}&\textbf{71.28}\\
			\hline
			ResNet56&filter level&94.33&72.29\\
			&layer level&\textbf{94.73}&\textbf{72.83}\\
			\hline
			ResNet110&filter level&94.09&74.24\\
			&layer level&\textbf{94.96}&\textbf{75.27}\\
			\hline
			MobileNetv2&filter level&92.66&72.70\\
			&layer level&\textbf{94.20}&\textbf{74.15}\\
			\hline
		\end{tabular}\\
	\end{center}
	\caption{Filter level grafting vs. layer level grafting}
	\label{table:filter level grafting}
\end{table}

\subsection{Two forms of  the Layer Information}\label{sec_s_4}

When calculating the layer information, we propose two forms (Equation (5) and Equation (6)) in the main paper. Equation (5) calculates the layer information as the sum of all the filter's information in a certain layer. But when two filters are identical in the same layer, one is redundant for the other.
Equation (5) merely sums all filters' information, which neglects the correlation among filters, while Equation (6) takes such correlation into consideration and perform entropy calculation on the whole layer.
We perform an experiment with different entropy calculations and results are listed in Table \ref{table:filter level entropy}.

\begin{table}[!h]
	\begin{center}
		\begin{tabular}{|c|c|c|c|} 
			\hline
			model&method&CIFAR-10&CIFAR-100\\
			\hline
			ResNet32&Equation (5)&93.89&70.95\\
			&Equation (6)&\textbf{93.94}&\textbf{71.28}\\
			\hline
			ResNet56&Equation (5)&94.40&72.03\\
			&Equation (6)&\textbf{94.73}&\textbf{72.83}\\
			\hline
			ResNet110&Equation (5)&94.48&74.34\\
			&Equation (6)&\textbf{94.96}&\textbf{75.27}\\
			\hline
			MobileNetv2&Equation (5)&93.41&72.86\\
			&Equation (6)&\textbf{94.20}&\textbf{74.15}\\
			\hline
		\end{tabular}\\
	\end{center}
	\caption{Different methods for calculating the layer information.}
	\label{table:filter level entropy}
\end{table}

From Table \ref{table:filter level entropy}, compared with Equation (5), Equation (6) shows more appealing performance improvements and is thus a better way to calculate the layer information.

\subsection{Efficiency of Adaptive Weighting Strategy} \label{sec_s_5}

We perform an experiment to compare adaptive weighting and fixed weighting strategies in Table \ref{table:adaptive_fixed}. For fixed weighting, $\alpha$ is fixed to be 0.5 in (2) from the main paper. From Table \ref{table:adaptive_fixed}, adaptive weighting performs better on each dataset and network structure, which proves the efficiency of adaptive strategy.

\begin{table}[!h]
	\small
	\begin{center}
		\begin{tabular}{|c|c|c|c|} 
			\hline
			model&method&CIFAR-10&CIFAR-100\\
			\hline
			ResNet32&fixed weighting&93.22&70.70\\
			&adaptive weighting&\textbf{93.94}&\textbf{71.28}\\
			\hline
			ResNet56&fixed weighting&94.54&72.25\\
			&adaptive weighting&\textbf{94.73}&\textbf{72.83}\\
			\hline
			ResNet110&fixed weighting&94.21&73.88\\
			&adaptive weighting&\textbf{94.96}&\textbf{75.27}\\
			\hline
			MobileNetv2&fixed weighting&93.48&73.52\\
			&adaptive weighting&\textbf{94.20}&\textbf{74.15}\\
			\hline
		\end{tabular}\\
	\end{center}
	\caption{Comparison of adaptive weighting and fixed weighting.}
	\label{table:adaptive_fixed}
\end{table}

\subsection{Proof of Theorem}\label{sec_s_1}
Here we prove the Theorem \ref{theo1} from Section \ref{sec_3_1_2}.

\begin{Proof}
	We first prove $H(Z|X) = H(Y|X)$:
	\begin{small}
		\begin{align*}
		&H(Z|X) \\
		&= -\sum_{x}p(X=x)\sum_{z}p(Z=z|X=x)\log P(Z=z|X=x)\\
		&= -\sum_{x}p(X=x)\sum_{z}p(Y=z-x|X=x) \log P(Y=z-x|X=x)\\
		&= -\sum_{x}p(X=x)\sum_{y}p(Y=y|X=x)\log P(Y=y|X=x)\\
		& = H(Y|X)
		\end{align*}
	\end{small}
	Then, according to the principle of entropy:
	\begin{align*}
	H(X,Y)& = H(X) + H(Y|X) \\
	&= H(X) + H(Z|X)\\
	&= H(X,Z)\\
	\end{align*}
	By symmetry of  entropy, the other direction also holds. Thus:
	\[
	H(X,Y) = H(X,Z) = H(Y,Z)
	\]
\end{Proof}

\end{document}